\documentclass[runningheads]{llncs}
\usepackage[T1]{fontenc}

\usepackage{amsmath,amssymb,amsfonts}
\usepackage{multirow}
\usepackage{booktabs}
\usepackage{subcaption}
\usepackage{marvosym}

\usepackage{graphicx,verbatim}
\begin{document}
\title{ReportMedSAM: Guiding Segmentation Through Radiology Reports}

\author{Anghong Du\inst{1}
\and Theodoros N. Arvanitis\inst{1}
\and Colin Watts\inst{2}
\and Alejandro F. Frangi,\inst{3,4}
\and Le Zhang\inst{1}\textsuperscript{(\Letter)}
}
\authorrunning{J. Liu et al.}
\institute{School of Engineering, College of Engineering and Physical Sciences,\\ University of Birmingham, Birmingham, UK \\
\and Department of Cancer and Genomic Sciences, College of Medicine and Health, University of Birmingham, Birmingham, UK\\
\and Department of Computer Science, Faculty of Science and Engineering, University of Manchester, Manchester, UK\\
\and The Division of Informatics, Imaging and Data Science, Faculty of Biology, Medicine, and Health, University of Manchester, Manchester, UK\\
     \email{axd1038@student.bham.ac.uk; l.zhang.16@bham.ac.uk}
}
  
\maketitle              
\begin{abstract}
Free-form radiology reports contain rich clinical descriptions, yet converting them for reliable segmentation remains challenging due to the inherent variability of natural language. Existing pipelines often rely on predefined organ phrases or brittle rule-based inference-time extraction, which limits their scalability to novel anatomical structures and makes them sensitive to linguistic variations. To address this, we propose ReportMedSAM, a report-driven framework that replaces discrete extraction with a learnable concept bank. By leveraging a frozen medical vision-language encoder (BiomedCLIP), we align organ-level concept embeddings with large-scale clinical corpora through contrastive learning, establishing mutually orthogonal semantic anchors. Our approach explicitly mitigates organ-level semantic collapse and ensures high robustness against diverse clinical synonyms (e.g., ``renal'' vs.\ ``kidney'').
During inference, a clinical report is embedded and matched against this concept bank to dynamically activate task-specific Mixture-of-Experts (MoE) modules. 
This decoupled design allows new concepts and experts to be added without retraining existing components, providing a parameter-isolated extension mechanism while keeping previously learned experts unchanged.
Evaluated on the AbdomenAtlas~3.0 dataset, ReportMedSAM effectively interprets free-form reports, achieves competitive segmentation accuracy, and demonstrates seamless, non-interfering extension to novel clinical tasks. \textit{Code will be released after review.}

\keywords{Segmentation  \and Foundation Model \and Mixture-of-Experts.}

\end{abstract}

\section{Introduction}

Recent advances in foundation models \cite{sam} \cite{medsam} have substantially improved the generalization capability of medical image segmentation across datasets and anatomical structures. Despite this progress, most existing segmentation frameworks still assume that segmentation targets are explicitly specified, either through predefined organ classes or user-provided prompts \cite{biomedparse} \cite{sam3} \cite{imis}. This assumption limits their applicability in real clinical workflows, where segmentation targets are rarely defined in advance. In routine clinical practice, medical images are typically accompanied by free-form radiology reports \cite{mimic} \cite{bassi2025radgpt} describing findings across multiple anatomical regions. These reports are unstructured, linguistically diverse, and often contain heterogeneous information that may or may not be directly relevant \cite{learningreport} to downstream segmentation. Consequently, determining \emph{which anatomical structures are mentioned and should be considered for segmentation} from such reports remains a fundamental challenge, particularly when multiple organs are described within a single narrative.

Recent studies have explored text-guided and language-assisted segmentation by leveraging vision-language models to associate textual descriptions with visual representations \cite{clip} \cite{biomedclip}. While these approaches reduce the reliance on dense annotations and enable flexible interaction, most existing methods specify segmentation targets using discrete phrases or predefined textual queries (e.g., ``segment the liver'') \cite{radgraph} \cite{cTAKES} \cite{chexpert}. Such phrase-based formulations are inherently sensitive to linguistic variation and do not generalize well to synonymous or clinically diverse expressions commonly observed in radiology reports. Furthermore, while emergent Large language Models (LLMs) offer sophisticated reasoning capabilities for interpreting clinical text \cite{llavamed} \cite{Med-PaLM}, 
stacking a multi-billion parameter LLM backbone atop a segmentation foundation model \cite{lora} \cite{switch} imposes high memory and latency overhead, creating significant barriers for real-time clinical workflows. More critically, in medical imaging scenarios, organ-level prompts often suffer from semantic collapse \cite{linguistic} \cite{biomedclip} \cite{medclip} in the embedding space: anatomically adjacent organs (e.g., liver and pancreas in abdominal CT) frequently co-occur in both images and reports, and are described using long-tailed, overlapping clinical terminology. As a result, their embeddings become overly clustered in the manifold \cite{does} \cite{AttTok}, weakening the fine-grained organ discrimination required for precision segmentation and reducing it to coarse region-level distinction.

\textbf{Our Contribution:} We present \textit{ReportMedSAM}, a report-driven medical image segmentation framework that directly addresses key limitations of existing approaches by enabling: \textbf{(1)} segmentation conditioned on free-form radiology reports, \textbf{(2)} automatic identification of anatomically relevant structures without relying on handcrafted templates or predefined queries, and 
\textbf{(3)} incremental adaptation to previously unseen anatomical targets through parameter-decoupled extensions. 
To achieve this, our framework decouples semantic target identification from pixel-level segmentation and learns organ-level concept anchors within a frozen BiomedCLIP embedding space. These trainable concept embeddings are aligned with large-scale clinical reports and regularized via inter-concept orthogonality, promoting stable semantic discrimination while keeping the text encoder fixed, providing robustness to diverse clinical synonyms.
During inference, targets are identified via similarity-based semantic routing. The parameter-isolated expert design enables computationally efficient integration of new anatomical structures without modifying previously learned components, making the framework naturally compatible with incremental extension.

\section{Method}

\begin{figure}[!t]
    \centering
    \includegraphics[width=\linewidth]{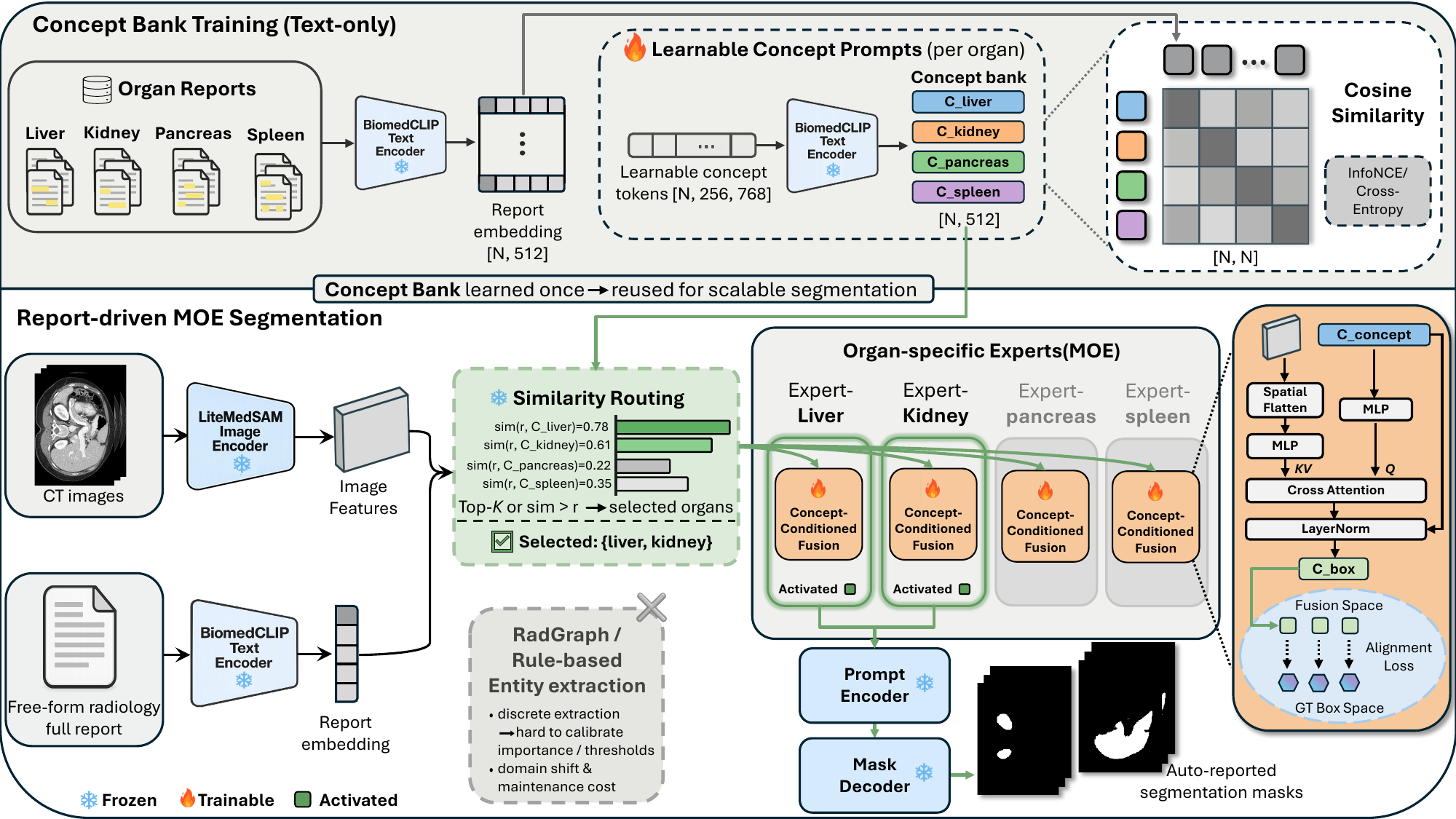}
    \caption{Overview of our proposed segmentation framework.}
    \label{fig:model_pipeline}
\end{figure}

\subsection{Problem Set-up}

We study the problem of \textbf{report-driven organ segmentation}, where segmentation targets are not explicitly specified but inferred from a free-form radiology report. Formally, let $\mathcal{D} = \{ (V_n, r_n) \}_{n=1}^{N}$ denote a dataset of $N$ cases. Each 3D volume $V_n$ is decomposed into 2D axial slices $\{x_{n,i}\}_{i=1}^{L_n}$, where $x_{n,i} \in \mathbb{R}^{H \times W \times C}$ represents the $i$-th input slice and $L_n$ is the number of slices in the volume. Here, $r_n$ is the report describing the entire volume $V_n$. During training, each slice is provided with ground-truth organ-wise binary masks $\mathbf{y}_{n,i} = \{ y_{n,i}^{(k)} \in \{0,1\}^{H \times W} \}_{k=1}^{K}$ for $K$ candidate anatomical structures.
Unlike conventional prompt-based segmentation where targets are explicitly cued, the subset of target organs mentioned in the report is not given at inference. We denote by $\mathcal{S}_n \subseteq \{1,\dots,K\}$ the \textbf{report-level organ set} inferred from $r_n$, which may contain complex linguistic phenomena such as negations or implicit references. For a volume decomposed into slices $\{x_{n,i}\}$, the task is formulated as estimating
$p(\{y_{n,i}^{(k)}\}_{k \in \mathcal{S}_n} \mid x_{n,i}, r_n)$.
A key challenge is the \textbf{spatial sparsity} of anatomical structures: an organ $k \in \mathcal{S}_n$ may appear in only a subset of slices, so $y_{n,i}^{(k)}$ may result in an all-zero mask for specific slices $x_{n,i}$ where the organ is physically absent.
Under this target-agnostic setting, the task requires both \textbf{semantic selection} (inferring $\mathcal{S}_n$ from $r_n$) and \textbf{existence perception} (determining the presence of $k$ in $x_{n,i}$). 
Figure~\ref{fig:model_pipeline} illustrates the framework.

\subsection{Text-driven Target Inference via Organ Concepts}
In this section, we derive organ-specific report collections from raw clinical texts and construct organ-level textual concepts to serve as a semantic interface for report-conditioned target inference. We learn a set of trainable concept embeddings within a frozen BiomedCLIP text encoder, effectively bridging the gap between unstructured clinical narratives and precise segmentation targets.

\textbf{Organ-specific report sets.}
Let $\mathcal{R}=\{r_{n}\}_{n=1}^{N}$ denote the set of full radiology reports available during training. 
To bootstrap organ-level concept learning, we construct organ-associated report subsets using lightweight rule-based operators $g_{k}(\cdot)$, implemented via keyword matching and regular expressions. 
These operators are used only during concept pretraining to provide coarse semantic grouping and are not involved in inference-time target identification.
For each organ $k \in \{1,\dots,K\}$, the operator extracts organ-related segments $\tilde{r}_{n}^{(k)} = g_{k}(r_{n})$. 
This defines the filtered collection
$
\mathcal{R}_{k}
=
\left\{
\tilde{r}_{n}^{(k)}
\;\middle|\;
\tilde{r}_{n}^{(k)} \neq \emptyset,\; n = 1,\dots,N
\right\},
$
which aggregates diverse clinical descriptions of organ $k$ across the cohort. 
Due to the multi-organ nature of radiological findings, a single report may contribute to multiple such collections.

\textbf{Trainable Organ-Specific Concept Bank.}
We define a trainable Concept Bank $\mathcal{C}=\{c_{k} \in \mathbb{R}^{d}\}_{k=1}^{K}$ where each $c_{k}$ serves as a semantic prototype for organ $k$. These concepts are optimized as free parameters within the frozen medical text embedding space rather than adhering to fixed labels. Using a frozen BiomedCLIP text encoder $f_{text}(\cdot)$, each report excerpt is mapped to an embedding $z_{n}^{(k)}=f_{text}(\tilde{r}_{n}^{(k)})$. The bank is optimized via contrastive learning to align report embeddings with their corresponding concepts while remaining distinct from others. To further ensure semantic decoupling and mitigate redundancy, we enforce an inter-concept orthogonality constraint. This forces the concepts to span near-orthogonal directions, establishing unique semantic anchors that capture organ-specific information invariant to synonyms.

\textbf{Report Target Inference.} Anatomical targets are identified from the full report embedding $z_n = f_{\text{text}}(r_n)$ by computing cosine similarities $s_n^{(k)} = \cos(z_n, c_k)$ for $k \in \{1, \dots, K\}$. The target set $\mathcal{S}_n$ is determined via a selection operator $\sigma$:
\begin{equation}
\mathcal{S}_n = \sigma \left( \{s_n^{(k)}\}_{k=1}^K ; \phi \right)
\end{equation}
where $\phi$ denotes the routing configuration. Specifically, $\sigma$ is implemented as either thresholding $\{k \mid s_n^{(k)} > \tau\}$ or Top-$K'$ ranking $\operatorname{arg\,top}_{K'} \{s_n^{(k)}\}_{k=1}^K$, which selects the $K'$ most relevant anatomical structures. These similarity-based signals serve as semantic gates to activate organ-specific segmentation experts, enabling the model to ground unstructured narratives into a precise anatomical focus.

\subsection{Concept-Conditioned Organ-wise Segmentation}
Given an input image $x_n$ and inferred target organ $k \in \mathcal{S}_n$ with its corresponding concept embedding $c_k$, we predict an organ-wise binary segmentation mask $\hat{y}_n^{(k)}$. This is formulated as a conditional segmentation problem $p(y_{n}^{(k)}|x_{n},c_{k})$, where the organ concept $c_k$ serves as a semantic conditioning variable. We build upon the LiteMedSAM \cite{litemedsam} framework, keeping the backbone frozen to preserve pre-trained medical knowledge \cite{seem}. 
Dense features $F_{n} = E_{img}(x_{n}) \in \mathbb{R}^{H \times W \times d}$ are extracted and reshaped into spatial tokens 
$V_n \in \mathbb{R}^{HW \times d}$, 
where $HW$ denotes the number of locations (e.g., $HW=4096$). 
An organ-specific lightweight projection is then applied:
$V_n^{(k)} = MLP_k(V_n)$,
with $V_n^{(k)} \in \mathbb{R}^{HW \times d}$.
We map the visual–concept pair $(V_n^{(k)}, c_k)$ to a fused representation $h_n^{(k)} = \Phi_k(V_n^{(k)}, c_k)$ via cross-attention, where the concept embedding $c_k \in \mathbb{R}^{d}$ serves as the query and the spatial visual tokens $V_n^{(k)}$ serve as the key and value. 
Attention is computed over the $HW$ spatial tokens, enabling query-conditioned spatial aggregation of visual evidence.
The fused representation is then mapped to geometric prompts 
$(p_{sparse,n}^{(k)}, p_{dense,n}^{(k)}) = \Psi_k(h_n^{(k)})$
by a trainable generator $\Psi_k$. 
The final mask is predicted by the frozen decoder
$\hat{y}_{n}^{(k)} = D_{mask}(F_{n}, p_{sparse,n}^{(k)}, p_{dense,n}^{(k)})$.
As the decoder is shared across organs, specialization is induced through the synthesized prompts. 
During training, ground-truth boxes $b_n^{(k)}$ act as spatial priors to regularize $\Psi_k$, aligning $p_{sparse,n}^{(k)}$ with reference embeddings from the frozen prompt encoder. 
At inference, the model performs fully automatic segmentation using only $x_n$ and $c_k$, without manual geometric cues.

\subsection{Training Objectives and Optimization}
The training of ReportMedSAM is conducted in a decoupled, two-stage manner to ensure scalability and prevent catastrophic interference. In the first stage, the Concept Bank is optimized for semantic grounding, while the second stage focuses on report-conditioned visual segmentation.

\noindent\textbf{Stage I: Concept Bank Optimization (Text-only).}
In the initial stage, a set of trainable concept embeddings $\mathcal{C}=\{c_{k}\}_{k=1}^{K}$ is optimized within the frozen BiomedCLIP text embedding space. We employ contrastive learning \cite{contrastivelearning} to align these concepts with organ-specific report segments. Specifically, for each organ concept, report segments from other organs within the same mini-batch serve as negative samples to ensure discriminative grounding.
To ensure organ-level discrimination and mitigate semantic collapse, an inter-concept orthogonality constraint is enforced:
$
\mathcal{L}_{ortho}=\sum_{k\ne k^{\prime}}\cos(c_{k},c_{k^{\prime}})
$.
This allows the concepts to function as unique semantic anchors that are invariant to clinical synonyms.

\noindent\textbf{Stage II: Report-driven Segmentation Training.}
During the second stage, the LiteMedSAM image encoder and mask decoder remain frozen to preserve pre-trained medical knowledge. Only the organ-specific projection layers and prompt synthesis modules are optimized. The loss function comprises three components:

\textbf{Organ-wise Segmentation and Suppression Loss ($\mathcal{L}_{seg}^{(k)}$):} For each inferred target organ $k \in \mathcal{S}_n$, we apply a combination of Dice and Binary Cross-Entropy (BCE) losses. To handle slice-wise organ visibility and suppress hallucinations, we define a unified mask loss:
\begin{equation}
\mathcal{L}_{seg}^{(k)} = 
\begin{cases} 
\beta_{pos} \left( \mathcal{L}_{Dice}(\hat{y}_n^{(k)}, y_n^{(k)}) + 0.5\mathcal{L}_{BCE}(\hat{y}_n^{(k)}, y_n^{(k)}) \right), & \text{if } \sum y_n^{(k)} > 0 \\
\beta_{neg} \left( \mathcal{L}_{Dice}(\hat{y}_n^{(k)}, \mathbf{0}) + 0.3\mathcal{L}_{BCE}(\hat{y}_n^{(k)}, \mathbf{0}) \right), & \text{if } \sum y_n^{(k)} = 0
\end{cases}
\end{equation}
where $\mathbf{0}$ denotes an all-zero mask, set $\beta_{pos}=0.5$ and $\beta_{neg}=0.3$. This loss supervises positive detections when the organ is present and penalizes hallucinations for organs mentioned in the report but absent in the slice.
To further reduce false-positive activations in slices without any target organs, we additionally apply a \textbf{Background Suppression Loss ($\mathcal{L}_{bg}$)}. It employs the same Dice and BCE combination as the negative branch of $\mathcal{L}_{seg}^{(k)}$ to ensure the model remains silent when anatomical structures are absent.

\textbf{Prompt Alignment Regularization ($\mathcal{L}_{align}^{(k)}$):} 
To stabilize the mapping from semantic concepts to the prompt space, synthesized sparse embeddings $p_{sparse,n}^{(k)}$ are regularized against reference box embeddings $p_{box,n}^{(k)}$ produced by the frozen prompt encoder:
\begin{equation}
\mathcal{L}_{align}^{(k)} = (1-\cos(p_{sparse,n}^{(k)},p_{box,n}^{(k)})) + \gamma ( \|p_{sparse,n}^{(k)}\|_{2}-\|p_{box,n}^{(k)}\|_{2} )^{2}
\end{equation}
where $\gamma = 0.05$ prioritizes directional consistency in the embedding manifold.

\noindent\textbf{Overall Objective.} The final training objective for Stage II is formulated as a weighted combination of the aforementioned losses:
\begin{equation}
\mathcal{L} = \sum_{k \in \mathcal{S}_n} \left( \lambda_{seg} \mathcal{L}_{seg}^{(k)} + \lambda_{align} \mathcal{L}_{align}^{(k)} \right) + \lambda_{bg} \mathcal{L}_{bg}
\end{equation}

\section{Experiments}

\textbf{Dataset.}
We evaluate ReportMedSAM on the AbdomenAtlas 3.0 dataset (liver, pancreas, kidneys, spleen). To ensure data independence, we select 1,200 cases and split them at the case level into 1,081 for training (137k slices) and 115 for testing (14k slices). The training set is further divided into training/validation subsets (90\%/10\%). 
To suppress hallucinations,  5\% background slices are included in both subsets. 
Official masks and reports provide ground truth, with bounding boxes derived automatically from masks via tight enclosing boxes and organ-specific text distilled from reports via keyword matching.

\textbf{Implementation Settings.} 
ReportMedSAM was implemented in PyTorch using a frozen LiteMedSAM backbone and a frozen BiomedCLIP text encoder. Input images were resized to $256 \times 256$, min-max normalized to $[0,1]$, and augmented via random flipping and rotation. 512-dimensional text embeddings were $L_2$-normalized prior to routing. Optimization was performed using AdamW ($LR=10^{-3}$, weight decay $= 2\times10^{-5}$) for 100 epochs with a batch size of 96 on an NVIDIA A100 40GB GPU. For Stage I contrastive training, the batch size was 1024 and the temperature parameter was set to 0.07. The training objective utilized loss weights $\lambda_{seg}=1.0$, $\lambda_{box}=0.5$, and $\lambda_{bg}=0.1$, with negative supervision scaled by a factor of 0.3 to suppress hallucinations.
For the prompt-based evaluation, baseline models received raw phrases and  synonyms in strict accordance with their original configurations, without external term-mapping.

\textbf{Baseline Settings and Evaluation Metrics.} 
ReportMedSAM is compared against BioMedParse, SAM3, and IMIS-Net. We evaluate linguistic robustness by testing both typical anatomical terms and equivalent synonyms. 
Performance is quantified by Dice Score for spatial overlap, False Activation Rate (FAR) for erroneous expert activation when an organ is absent, and Precision for the purity of selected concepts.

\begin{table}[!t]
\fontsize{8}{9}\selectfont
    \setlength{\tabcolsep}{1mm}
    \renewcommand{\arraystretch}{0.9}
    \centering
    \caption{The results of Dice score on the AbdomenAtlas 3.0 dataset.
    Bold numbers denote the best performance among all methods.
    Statistical significance is assessed using paired $t$-tests on case-level Dice scores, where bold results indicate methods that are significantly better than others ($p < 0.01$).}
    \begin{tabular*}{\linewidth}{@{\extracolsep{\fill}}c|c|c|cccc|c}
       \toprule
        {\textbf{Prompt}} & {\textbf{Methods}} & {\textbf{Baseline}} & Liver & Spleen & Pancreas & Kidney & Avg \\ 
        \midrule

        \multirow[c]{4}{*}{Typical} & Concept & Ours & \textbf{0.810} & 0.681 & 0.396 & 0.707 & 0.647 \\ 
        \cmidrule{2-8} 
        & \multirow[c]{3}{*}{RadGraphXL} & BioMedParse & 0.582 & 0.638 & 0.399 & \textbf{0.844} & 0.615 \\ 
        & & SAM3 & 0.737 & \textbf{0.772} & \textbf{0.478} & 0.662 & \textbf{0.662} \\ 
        & & IMIS-Net & 0.388 & 0.000 & 0.011 & 0.313 & 0.178 \\
        
        \midrule 

        {\textbf{Prompt}} & {\textbf{Methods}} & {\textbf{Baseline}} & 
        Hepatic & Splenic & Pancreatic & Renal & Avg\\ 
        \midrule
        
        \multirow[c]{4}{*}{Synonyms} & Concept & Ours & \textbf{0.810} & \textbf{0.681} & \textbf{0.396} & 0.707 & \textbf{0.647} \\ 
        \cmidrule{2-8}
        & \multirow[c]{3}{*}{RadGraphXL} & BioMedParse & 0.559 & 0.639 & 0.391 & \textbf{0.844} & 0.608 \\ 
        & & SAM3 & 0.165 & 0.000 & 0.078 & 0.003 & 0.003 \\ 
        & & IMIS-Net & 0.017 & 0.004 & 0.045 & 0.142  & 0.052 \\
    
        \bottomrule
    \end{tabular*}
    \label{tab:AbdomenAtlas}
\end{table}

\textbf{Comparison with SOTA Methods.}
Table~\ref{tab:AbdomenAtlas} shows ReportMedSAM maintains a stable 0.647 average Dice across both linguistic settings, whereas baselines like SAM3 and IMIS-Net fail under synonyms due to rigid text matching. This resilience stems from leveraging the pre-trained BiomedCLIP manifold to preserve semantic proximity between clinical synonyms (e.g., ``liver'' and ``hepatic''). Consequently, even for unseen clinical expressions, ReportMedSAM leverages stable similarity gradients to map variable inputs to fixed, orthogonal concept anchors, effectively resolving the linguistic variability inherent in radiology reports. 
Furthermore, our results were obtained via a sequential incremental training protocol with frozen experts, empirically validating the framework's scalability.

\begin{table*}[!t]
\fontsize{8}{9}\selectfont
    \setlength{\tabcolsep}{1mm}
    \renewcommand{\arraystretch}{0.9}
    \centering
    \caption{
    Top-$K$ routing performance. FAR ($\downarrow$) measures erroneous segmentation for absent organs; Precision ($\uparrow$) evaluates concept selection purity.
    }
    
    \begin{tabular*}{\linewidth}{@{\extracolsep{\fill}}c c cc cc cc cc}
        \toprule
        \multirow{2}{*}{\textbf{Methods}} 
        & \multirow{2}{*}{\textbf{$K$}} 
        & \multicolumn{2}{c}{\textbf{Liver}} 
        & \multicolumn{2}{c}{\textbf{Spleen}} 
        & \multicolumn{2}{c}{\textbf{Pancreas}} 
        & \multicolumn{2}{c}{\textbf{Kidney}} \\
        
        \cmidrule(lr){3-4}
        \cmidrule(lr){5-6}
        \cmidrule(lr){7-8}
        \cmidrule(lr){9-10}
        
        & & FAR$\downarrow$ & Prec$\uparrow$ 
          & FAR$\downarrow$ & Prec$\uparrow$
          & FAR$\downarrow$ & Prec$\uparrow$
          & FAR$\downarrow$ & Prec$\uparrow$ \\
        \midrule

        RadGraphXL & -- 
        & 1.000 & 0.750 
        & 1.000 & 0.750 
        & 1.000 & 0.750 
        & 1.000 & 0.750 \\ 
        
        \midrule
        
        \multirow{4}{*}{Concept}
        & 4 
        & 1.000 & 0.750 
        & 1.000 & 0.750
        & 1.000 & 0.750
        & 1.000 & 0.750 \\ 
        
        & 3 
        & 0.375 & 0.875 
        & 0.990 & 0.671
        & 0.220 & 0.927
        & 0.804 & 0.732 \\ 
        
        & 2 
        & 0.237 & 0.881
        & 0.758 & 0.621
        & 0.067 & 0.967
        & 0.694 & 0.653 \\ 
        
        & 1 
        & 0.207 & 0.793 
        & 0.539 & 0.461
        & 0.001 & 0.999
        & 0.601 & 0.400 \\ 
        
        \bottomrule
    \end{tabular*}
    \label{tab:topk}
\end{table*}

    

\begin{figure}[!t]
    \centering
    \includegraphics[width=\linewidth]{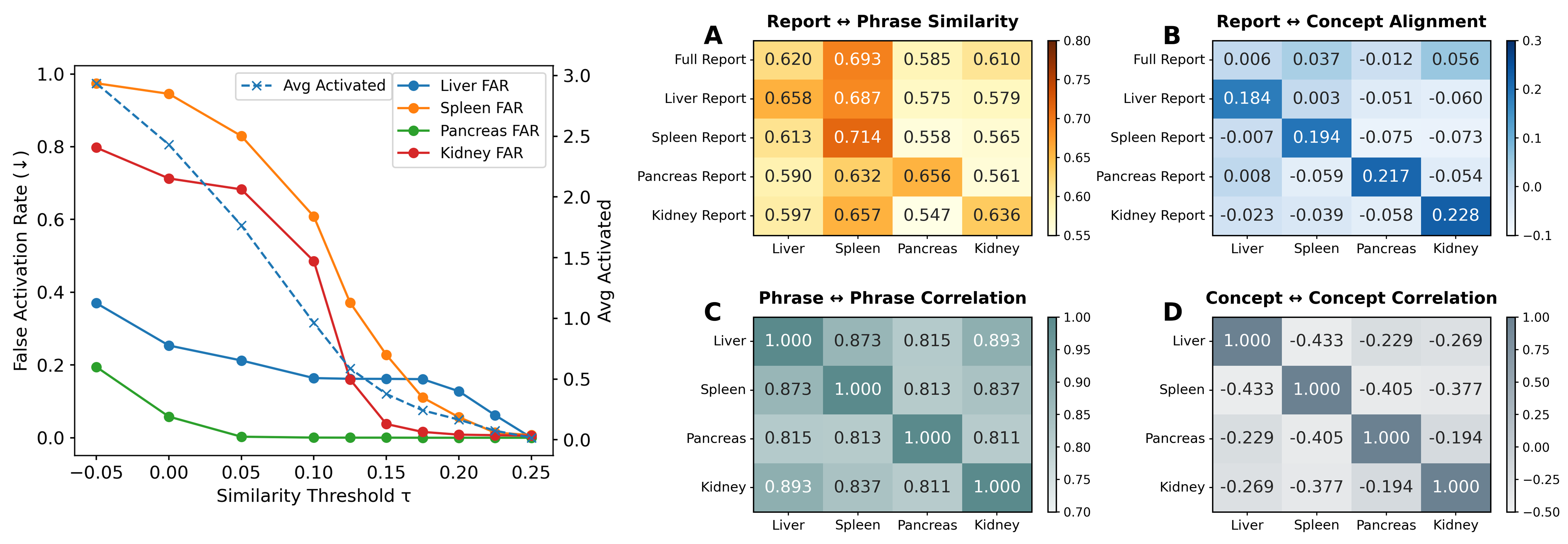}
    \caption{Quantitative analysis of report-driven routing and the learned semantic manifold. (Left) Effect of the similarity threshold $\tau$ on false activation. Solid lines denote organ-wise FAR, while the dashed line indicates the average number of activated organs. (Right) Semantic consistency analysis across representations.}
    \label{fig:zhexian_cos}
\end{figure}

\textbf{Ablation Experiment.}
We investigate the mechanisms mitigating semantic collapse and ensuring reliable report-driven routing. Anatomically adjacent organs frequently co-occur in clinical narratives and share overlapping terminology, causing standard phrase-level representations to exhibit high cross-correlation (e.g., 0.873 between liver and spleen in Fig.~\ref{fig:zhexian_cos} (right, C). By optimizing our \textbf{Concept Bank} within the frozen \textbf{BiomedCLIP} space using $\mathcal{L}_{ortho}$, we establish a \textbf{near-orthogonal concept geometry} that significantly reduces cross-organ interference (Fig.~\ref{fig:zhexian_cos} (right, D)).
This preserves synonym proximity (e.g., ``liver'' vs. ``hepatic'') while establishing unique, linguistic-invariant anchors that resolve entanglement even for long-tailed clinical expressions. 
Regarding routing reliability, we evaluate two proposed strategies: \textbf{Top-$K$ ranking} and \textbf{similarity thresholding}. 
Table~\ref{tab:topk} shows that rule-based extraction (e.g., RadGraphXL) yields a False Activation Rate of $1.000$. While NLP tools can identify anatomical entities and their status, they operate at the entity level without similarity-based prioritization, resulting in high false activation rates under our routing-based evaluation protocol.
In contrast, our Top-$K$ strategy provides a discrete selection interface. As shown in Table~\ref{tab:topk}, a $K=1$ strategy suppresses false activations to an average FAR of $0.336$ while maintaining high precision for specific experts. Alternatively, the thresholding strategy in Fig.~\ref{fig:zhexian_cos} (Left) offers a continuous, calibratable interface. Increasing the similarity threshold $\tau$ from $-0.05$ to $0.25$ reduces the average number of activated organs from $3.0$ to $0.0$ , enabling a controllable trade-off between segmentation recall and routing precision. These results indicate that the learned concept anchors effectively bridge the gap between clinical text and segmentation intent through flexible routing mechanisms.



\section{Conclusion}

Free-form radiology reports and medical images are inherently paired in routine clinical workflows and harbor extensive anatomical information, yet remain underutilized for segmentation, as conventional methods rely on explicit targets or handcrafted prompts. We present ReportMedSAM, a framework that infers segmentation targets from clinical text via similarity-based routing, eliminating manual prompting or rule-based selection at inference.
To resolve semantic ambiguities from anatomical proximity and linguistic diversity, ReportMedSAM establishes organ-level anchors within a frozen medical vision-language space. Its parameter-decoupled architecture activates organ-specific experts, facilitating modular expansion to novel structures without affecting existing components. 
Experiments demonstrate competitive multi-organ segmentation performance and highlight its potential as a practical semantic interface for foundation model--based medical image segmentation.

\bibliographystyle{splncs04}  
\bibliography{Paper-0555}    

\end{document}